\documentclass[conference,a4paper,10pt ]{IEEEtran}
\IEEEoverridecommandlockouts
\usepackage{cite}
\usepackage{amsmath,amssymb,amsfonts}
\usepackage{algorithmic}
\usepackage{graphicx}
\usepackage{textcomp}
\usepackage{xcolor}

\def\BibTeX{{\rm B\kern-.05em{\sc i\kern-.025em b}\kern-.08em
    T\kern-.1667em\lower.7ex\hbox{E}\kern-.125emX}}

\usepackage{amsmath,amssymb,amsfonts, enumitem, fancyhdr, color, comment, graphicx, environ,changepage, mathrsfs}
\usepackage{tikz-cd}
\usepackage{quiver}
\usepackage{hyperref}
\usepackage[inkscapelatex=false]{svg}
\usepackage{caption}
\usepackage{subcaption}
\usepackage{bbm}
\usepackage{soul}

\def\b{\mathbf{b}}
\def\x{\mathbf{x}}
\def\w{\mathbf{w}}

\def\B{\mathbf{B}}
\def\T{\mathbf{T}}

\newcommand{\btheta}{\boldsymbol\theta}
\newcommand{\bTheta}{\boldsymbol\Theta}

\newcommand\pearl[1]{{\widetilde{#1}}}

\newcommand{\C}{{\bf C}}
\newcommand{\bzeros}{\boldsymbol{0}}

\begin{document}

\title{On Counterfactual Interventions in Vector Autoregressive Models
\thanks{This work was supported by the National Science Foundation under Awards 2021002 and 2212506.}
}

\author{
\IEEEauthorblockN{Kurt Butler, Marija Iloska, and Petar M. Djuri\'{c}}
\IEEEauthorblockA{\textit{Department of Electrical and Computer Engineering},
Stony Brook University,
Stony Brook, NY 11794, USA
\\
\small{\texttt{kurt.butler@stonybrook.edu; marija.iloska@stonybrook.edu; petar.djuric@stonybrook.edu}}}
}

\maketitle

\begin{abstract}
Counterfactual reasoning allows us to explore hypothetical scenarios in order to explain the impacts of our decisions. However, addressing such inquires  is impossible without establishing the appropriate mathematical framework. In this work, we introduce the problem of counterfactual reasoning in the context of vector autoregressive (VAR) processes. We also formulate the inference of a causal model as a joint regression task where for inference we use both data with and without interventions. After learning the model, we exploit linearity of the VAR model to make exact predictions about the effects of counterfactual interventions. Furthermore, we quantify the total causal effects of past counterfactual interventions.
The source code for this project is freely available at \url{https://github.com/KurtButler/counterfactual_interventions}.
\end{abstract}


\section{Introduction}
In many situations, we may be interested in ``what-if'' questions. As decisions in real life are rarely made with perfect knowledge of their effects, it is only natural that later one may question the optimality of their choices and ask how things would be if a different decision had been made. 
This process of reasoning about decisions and actions that contradict what actually occurred pertains to problems of counterfactual reasoning.

A counterfactual event is an event which does not agree with a particular outcome that was observed in a given experiment \cite[p. 29]{pearl2009causality}. For example, a student applying to a college might be rejected due to  a low score on an entrance exam. A counterfactual question could then be ``would this student have been accepted if their score were higher?'' In asking this question, we are imagining a hypothetical scenario in which we modify the score of this student, and keep all other variables (i.e., the scores of all the other applicants) constant. In general, counterfactual questions may help us understand whether our actions are effective. 

For the problem of detecting the presence of causal relationships and learning the graphical structure of a causal system, a large variety of approaches have been proposed, often arising from different fundamental principles \cite{rubin1974estimating,spirtes2000causation,zheng2018dags,schreiber2000measuring}. Many of these methods attempt to rely on entirely \emph{observational} data, where the experimenter cannot \emph{intervene} upon the system under study to probe into its cause and effect relationships. While it is sometimes possible to detect causalities without intervention, observational approaches often require strong assumptions that are difficult to verify empirically or are otherwise restrictive.For instance, identification of a linear causal model from only observations is ill-posed if the noises are Gaussian \cite[pp.50-51]{peters2017elements}. 
Methods based on predictability, such as Granger causality \cite{granger1969investigating}, can deliver false positives when not all relevant variables are included in the model. 
Other methods may require strong assumptions; %
for example, convergent cross mapping \cite{sugihara2012detecting} requires the existence of a dynamical attractor, which is difficult to test on small data sets \cite{butler2023causal}. In comparison, when it is possible to perform interventions, causal inference becomes considerably simpler, with several standard algorithms \cite{spirtes2000causation},\cite{ogarrio2016hybrid}. Under ideal conditions, when experiments can be carried out perfectly and there are no hidden processes, all causal relationships in system of $N$ variables can be detected with $N$ single-variable interventions \cite{eberhardt2006n}.

To answer counterfactual questions, more information than just the direction of causalities is required: one must also have a functional model of how the effect is determined by all of its causes \cite{pearl2009causality}.
For the college student example, we would need to know the entrance exam  scores of the other applicants to determine the functional model for acceptance and subsequently if the given student would be accepted, as their acceptance would depend on all of the scores. %

In this work, we consider the counterfactual prediction problem: Given a particular realization of a causal system, i.e., a multivariate time series and a causal model that explains it, can we predict what we would have observed if a previous, hypothetical intervention occurred?

We present several contributions relevant to studying counterfactual events with time series. We introduce an approach to learning the causal structure of a system by leveraging multiple data sets with different interventions. We apply this approach to the setting of vector autoregressive (VAR) models. After learning the causal model, we address the problem of making counterfactual predictions. Due to linearity of the VAR model, we are able to make exact predictions without needing to estimate unobserved variables. Finally, we validate our approach using several examples.

\section{Problem Formulation}
We consider the problem of modeling causal relationships from a multivariate time series $\x_t=[x_{1,t}, ..., x_{D,t}]^\top$. For each node $i$ and time instant $t$, we assume that the value of $x_{i,t}$ is determined by a function of the values of the system at previous time steps, $\x_{t-1},...,\x_{t-Q}$, up to some order $Q$, as well as a separate independent noise term $w_{i,t}$. As a simple example, we examine the vector autoregressive (VAR) model:
\begin{equation}
    \label{eq:ARmodel}
\x_t := \sum_{q=1}^Q \B_q \x_{t-q} + \w_t,
\end{equation}
where the $\B_q$ are matrices of linear coefficients, and $\w_t \sim \mathcal{N}(\mathbf{0},\mathbf{C})$, for some given noise covariance $\mathbf{C}$. The notation $:=$ expresses the direction of causality, where we consider \eqref{eq:ARmodel} as a causal generative process.

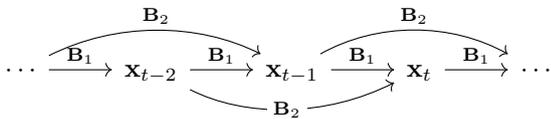
\begin{figure}[tbh]
    \centering
\[\begin{tikzcd}
	\cdots & {\x_{t-2}} & {\x_{t-1}} & {\x_t} & \cdots
	\arrow["{\B_1}", from=1-1, to=1-2]
	\arrow["{\B_1}", from=1-2, to=1-3]
	\arrow["{\B_1}", from=1-3, to=1-4]
	\arrow["{\B_1}", from=1-4, to=1-5]
	\arrow["{\B_2}", curve={height=-18pt}, from=1-1, to=1-3]
	\arrow["{\B_2}"{description}, curve={height=18pt}, from=1-2, to=1-4]
	\arrow["{\B_2}", curve={height=-18pt}, from=1-3, to=1-5]
\end{tikzcd}\]
    \caption{A causal graph obtained from \emph{unfolding} the causal relationships in \eqref{eq:ARmodel} to show their dependencies in time, in the $Q=2$ case. Each vertex represents a random vector, and each edge represents the presence of a cause-effect relationship. 
    }
    \label{fig:graph}
\end{figure}

According to \eqref{eq:ARmodel}, the components of $\x_{t-1}$, $\x_{t-2}$, and $\x_{t-Q}$ exert causal influences on $\x_t$. It is traditional in causality to express these relationships in terms of a graph, although there are several options as to how this can be done. In Figure \ref{fig:graph}, we present one approach that treats each time instance as a separate vertex, with the purpose of understanding the temporal dependencies of the VAR model, which will later simplify the calculation and prediction of counterfactuals.

To answer counterfactual queries about a system, we must first have a \textbf{structural causal model} (SCM) of the system \cite{pearl2009causality}. A SCM is a hypothesis about the generative process that created a data set. The SCM represents each variable in the system as a function of its parents, 
a subset of variables directly influencing a given variable,
along with an exogenous noise variable. 
In this way, the VAR model in \eqref{eq:ARmodel} can be interpreted as a SCM after we declare that the equation explicitly describes how the value of $\x_t$ is determined. 

Asserting that a VAR model is a causal model is a stronger statement than asserting it is a statistical model. A statistical model needs only to generate the correct joint probability distribution over the family of variables, but a causal model is a hypothesis about the generative process as it exists in reality. 
In the same way that multiple generative processes can represent the same probability distribution, there might be multiple SCMs that are compatible with a given data set. Additionally, different causal models will produce different answers to counterfactual questions, so care should be taken to learn the most accurate causal model before discussing counterfactuals.
In the next section, we introduce interventions and how SCMs respond to them.

\subsection{Interventions}
A major feature of causal models is that they are modularized, meaning that we may think of them as a set of individual components which can be changed or modified. In our example, each variable $x_{i,t}$ receives a value according to the formula
\begin{equation}
\label{eq:mechanism}
x_{i,t} := \sum_{q=1}^Q \b_{q,i}^\top \x_{t-q} + w_{i,t},
\end{equation}
where $\b_{q,i}^\top$ is the $i$-th row of $\B_q$, and $w_{i,t}$ is the $i$-th element of ${\bf w}_t$ in \eqref{eq:ARmodel}. An \textbf{intervention at node $i$ and time $t$} is a modification to the causal model which replaces the function that assigns a value to $x_{i,t}$ with a new one. For example, we may intervene by replacing the function in \eqref{eq:mechanism} with \st{a new one}:
\begin{equation}
\label{eq:modifiedmechanism}
x_{i,t} := \sum_{q=1}^Q \pearl{\b}_{q,i}^\top \x_{t-q} + \widetilde{\sigma} w_{i,t}, + \pearl{u}_{i,t},
\end{equation}
where $\pearl{\b}_{q,i}$ are modified linear coefficients, $ \pearl{u}_{i,t}$ is a signal injected by the experimenter, and $\widetilde{\sigma}\in [0, 1]$ is a parameter that may be used to control the level of noise. Under ideal conditions, the experimenter's intervention perfectly controls the value of $x_{i,t}$ by tuning the parameters in \eqref{eq:modifiedmechanism} to achieve a desired effect. 

\begin{figure*}[!ht]
    \centering
    \includegraphics[width=0.95\textwidth]{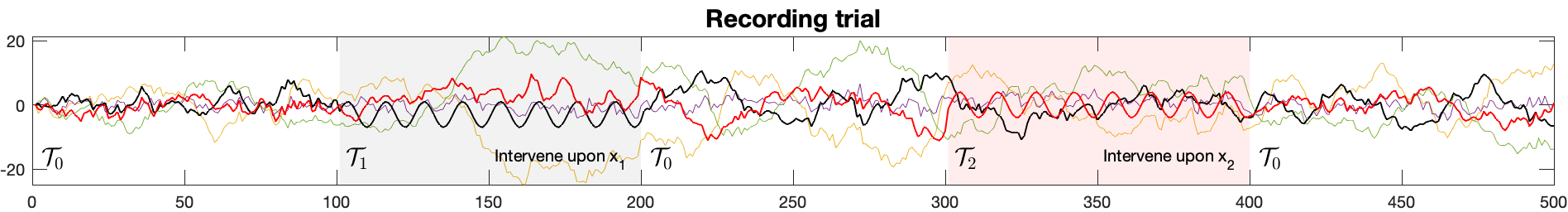}
    \caption{Interventions applied to a multivariate time series. The underlying causal model is a VAR model, \eqref{eq:ARmodel}, with $Q=2$ and $D=5$. Two separate interventions were performed during the experiments in which we drove a node in the system with a sinusoidal stimulus. For $t=101,...,200$, $x_{1,t}$ (black curve) was forced to take on the value $4\sin(t/2)$, and for $t=301,...,400$, $x_{2,t}$ (red curve) was set to $4\sin(t/2)$. After recording, we partitioned the trial into three data sets, ${\cal T}_0, {\cal T}_1, {\cal T}_2$, according to the intervention periods. Explicitly, ${\cal T}_1 = \{101, ..., 200\}$, ${\cal T}_2 = \{301, ..., 400\}$, and ${\cal T}_0 = \{1, ..., 500\} \setminus ({\cal T}_1 \cup {\cal T}_2)$ }.
    \label{fig:topfig}
\end{figure*}

In this paper, we assume that all interventions are able to perfectly control the value of the node intervened upon, so that %
$x_{i,t}$ satisfies \eqref{eq:modifiedmechanism} whenever an intervention on $x_{i,t}$ is being performed.
While intervening on a node $i$, it is assumed that the functions governing other nodes in the system remain unchanged, which is a form of modularity of the system. In real life, there might be some uncertainty in the values of $\pearl{\b}_{q,i}$ and $\pearl{\sigma_i}$ when the intervention is performed, due to imperfect experimental conditions.
An intervention can be applied for a period of time, during which we can actively observe the behavior of the rest of the system, i.e., the unperturbed nodes. We illustrate this process visually in Figure \ref{fig:topfig}.

Data that we record while performing an intervention are called \emph{interventional}, contrasting with \emph{observational} data that we record while not altering the system under study. 
While performing an experiment with multiple interventions, we allocate recorded data into separate data sets according to the type of intervention being applied (cf. Figure \ref{fig:topfig}). 
In Section \ref{sec:inference}, we introduce a method to leverage multiple data sets to improve estimation of $\B_q$ over purely observational data. 
The caveat is that this approach is only possible when obtaining interventional data is feasible.

\subsection{The problem of predicting counterfactuals}
Having introduced SCMs and interventions, we may now formulate \textbf{counterfactuals} in our framework. 
In short, a counterfactual is a hypothetical intervention. It is not an intervention that occurred in reality, but rather one whose behavior we can predict, %
having already observed a realization of the system. 

Now that the necessary causal background has been introduced, we can state the primary problem of this work. Consider an experiment in which we observe a multivariate signal $\x_t$ until a certain time $T$, and suppose that we have assumed a particular SCM for $\x_t$. Now assume we are interested in a hypothetical intervention that would have occurred in the past, and want to predict what we would have observed in this hypothetical scenario. That is, 
\begin{align}
    & \textbf{Given: }  \text{A particular realization } \{ \x_t | t=1,...,T\}, \notag \\  & \text{ and an SCM for $\x_t$},
    \\
    & \textbf{Predict: } \pearl{\x}_t, 
\end{align}
where  $\pearl{\x}_t$ are the signals in the case that we performed a hypothetical intervention upon $x_{i,t}$ for some $t$ in the range $1<T_1 \leq t \leq T_2 < T$, and for a fixed $i$. Here we make an important assumption that all conditions of the experiment, such as the particular realization of the noise process $\w_t$, remains the same in both cases. This is necessary, because in considering the counterfactual problem, asserting that all other conditions were the same implies that the noise too, should have the same realization. 

In general models, the values of $\w_t$ need to be estimated before counterfactual analysis can be performed. However, this estimation is straightforward in any additive noise model since $\w_t$ enters the formulas additively. In our approach, we exploit linearity of the system to circumvent the need to estimate $\w_t$ at all because the influence of the noise will cancel out. 

\section{Proposed Solution}

Before making counterfactual predictions, it is essential to first learn the causal structure of the system.
To learn the parameters of the causal model in a way that leverages both observational and interventional data, we propose a joint regression task that allows us to infer the parameters of each module shared across datasets. We make the tacit assumption that these interventions are ideal, and that the parameters of the model are unchanged when they are not related to the intervention.
We now introduce some notation. If at time $t$ we perform an intervention to control the value of $x_{i,t}$, then the time point $t$ should be allocated to a set of time indices ${\cal T}_i$.
If no intervention was performed, then the index $t$ of $\x_t$ is allocated to the index set ${\cal T}_0$. We assume that only one node can be intervened upon per time step $t$, so ${\cal T}_i \cap {\cal T}_j = \emptyset$ whenever $i \neq j$. We define 
\begin{equation}
{\cal T}_{-i} =  \left( \bigcup_{j=0}^D {\cal T}_j \right) \setminus {\cal T}_i
\end{equation}
to be the set of time indices of $x_i$, where $x_i$ was not subject to intervention. It is permissible that some index sets may be empty, ${\cal T}_j = \emptyset$.
In principle, these sets could be partitions of a single recording trial, or obtained from multiple recording trials, with no major modification to our approach.

\subsection{Causal structure learning}
\label{sec:inference}
For clarity in presentation, we will first describe the case of one lag $(Q=1)$, and from there we extend to the multiple lag case.

\paragraph{SINGLE LAG}
The goal is to learn the linear coefficient matrix $\B$ of the true causal model jointly across datasets by minimizing the following objective,
\begin{equation}
    \label{eq:naiveobjective}
    \min_{\B} \sum_{t \in \mathcal{T}_0} || \x_t-\B \x_{t-1}||_2^2 
+
\sum_{i=1}^D
\sum_{t \in \mathcal{T}_i} || \x_t^{[-i]}-\B^{[-i]} \x_{t-1}||_2^2
\end{equation}
where the operation $\mathbf{A}^{[-i]}$ removes row $i$ from the matrix $\mathbf{A}$.

Using the identity $||\x||^2_2 = x_1^2 + \cdots + x_D^2$, we can simplify this optimization task considerably. 
Recall that whenever we intervene on a particular variable, all other linear coefficients remain unchanged. Thus, the function that assigns a value to $x_{i,t}$ is given by 
$$
x_{i,t} := \b_i^\top \x_{t-1} + \w_t 
$$
whenever $t \notin {\cal T}_i$, or equivalently, when $t \in {\cal T}_{-i}$. These considerations lead us to propose the following objective, %
which is equivalent to \eqref{eq:naiveobjective}:
\begin{equation}
    \label{eq:stratifiedobjective}
    \min_{\B} \sum_{i=1}^D \sum_{t \in {\cal T}_{-i}} || x_{i,t} - \b_i^\top \x_{t-1} ||_2^2,
\end{equation}
where, again, $\b_i^\top$ is the $i$-th row of $\B$.

The revised objective in \eqref{eq:stratifiedobjective} has multiple advantages. Firstly, it can be solved for each variable $i$ separately as a distinct subproblem. Secondly, it makes it clear that minimum mean squared estimate of $\B$ can be obtained using classical methods. 
If we assume some level of sparsity in $\B$, then we can use LASSO\cite{tibshirani1996regression}, or $\ell_1$ regularization, to get a sparse estimate:
\begin{equation}
    \label{eq:sparseobjective}
    \min_{\B} \sum_{i=1}^D \sum_{t \in {\cal T}_{-i}} || x_{i,t} - \b_i^\top \x_{t-1} ||_2^2 + \lambda ||\B||_1,
\end{equation}
where $\lambda$ is a regularization parameter.

\paragraph{MULTIPLE LAGS} To generalize to the case in which $Q>1$, we only need to modify the approach slightly. Consider the model in \eqref{eq:ARmodel}. If we define 
\begin{equation}
    \bTheta = \begin{bmatrix}
        \B_1 & \cdots & \B_Q 
    \end{bmatrix},
\end{equation}
and
\begin{equation}
    \bar{\x}_t = \begin{bmatrix}
        \x_{t-1} & \cdots & \x_{t-Q} 
    \end{bmatrix},
\end{equation}
then we can express \eqref{eq:ARmodel} as
\begin{equation}
    \label{eq:multilagcompact}
    \x_t := \bTheta \bar{\x}_t + \w_t.
\end{equation}
Now we can formulate a similar optimization task as before, giving us the multiple lag version of \eqref{eq:naiveobjective}:
$$
\min_{\bTheta} \sum_{t \in \mathcal{T}_0} || \x_{t}-\bTheta \bar{\x}_t||_2^2 
+
\sum_{i=1}^D 
\sum_{t \in \mathcal{T}_i} || \x_t^{[-i]}-\bTheta^{[-i]} \bar{\x}_t||_2^2,
$$
which can again be split into tractable sub-problems. Considering sparsity, the multiple lag version of \eqref{eq:sparseobjective} is given to be
\begin{equation}
    \label{eq:sparseMULTILAGobjective}
    \min_{\bTheta} \sum_{i=1}^D \sum_{t \in {\cal T}_{-i}} || x_{i,t} - \btheta_i^\top \bar{\x}_{t} ||_2^2 + \lambda ||\bTheta||_1,
\end{equation}
where $\btheta_i^\top$ is the $i$-th row of $\bTheta$. The use of sparsity constraints becomes increasingly relevant when considering multiple lags, since strongly autocorrelated signals can have multiple distinct models with similar predictive power \cite{cui2024topology}.

\subsection{Estimation of total causal effects}
\label{sec:totaleffect}
We now consider the problem of estimating the effects of a past counterfactual intervention. Suppose that we have observed $\x_t$ up until some time $T$. For some $t<T$, consider an intervention that modifies $x_{i,t}$ to take on a new value $x_{i,t}^*$. If we define $\Delta x_{i,t} = x_{i,t}^* - x_{i,t}$, then we can reason about the hypothetical intervention as an additive change in $x_{i,t}$:
$$
x_{i,t} \longrightarrow x_{i,t} + \Delta x_{i,t}.
$$
Since the VAR model is linear, the additive change in $x_{i,t}$ induces an additive change in all variables ``downstream'' from it in the graph. For example, if we move one step into the future, then $\x_{t+1}$ changes according to 
$$
\x_{t+1} \longrightarrow \x_{t+1} + \B_1 \mathbf{e}_i \Delta x_{i,t},
$$
where $\mathbf{e}_i$ is the $i$-th unit vector in $\mathbb{R}^D$. We observe that the linear model coefficients in $\B_1$ measure the sensitivity of $\x_{t+1}$ to changes in $\x_t$, and can be interpreted as a measure of the strength of the causal relationship \cite{butler2022differential}. 

If we perform an intervention at time $t$ and attempt to predict its effect at time $t+k$, we need to propagate the changes through the graph in Figure \ref{fig:graph}, that is, through each intermediate time step. When $Q>1$, the path in the graph from $\x_t$ to $\x_{t+k}$ is not unique, and the total sensitivity of $\x_{t+k}$ to $\x_t$ is given by a sum over all paths in the graph \cite{liu2023detecting,salehkaleybar2020learning}. Since the VAR model is time-invariant, the total sensitivity is a matrix $\T_k$ which only depends upon $k$. We refer to this matrix as a \textbf{total causal effect} matrix. 

To compute the total causal effect matrix for the VAR model, we can exploit recursion. Suppose that $k > Q$. Any path from $\x_t$ to $\x_{t+k}$ must go through a node between $\x_{t+k-Q}$ and $\x_{t+k-1}$. As a result, the total causal effect decomposes according to the same rule:
\begin{equation}
    \T_k = \B_1 \T_{k-1} + \B_2 \T_{k-2} + \cdots + \B_Q \T_{k-Q}.
\end{equation}
If $k \leq Q$, then a similar rule holds, but we define $\T_0 = \mathbf{I}$ to be the identity matrix, and $\T_k =0$ for any $k<0$, since causation can only move forward in time.

\section{Experiments}
In this section, we demonstrate our approach to modeling counterfactuals with two
examples. First, we demonstrate how joint regression can be leveraged to more accurately learn the causal model. Second, we explore an example in which we study the effects of a past intervention on future, forecasted events. The presented results are of single outcomes of the experiments. 

\begin{figure}[b!]
    \centering
    \includegraphics[width=0.45\textwidth]{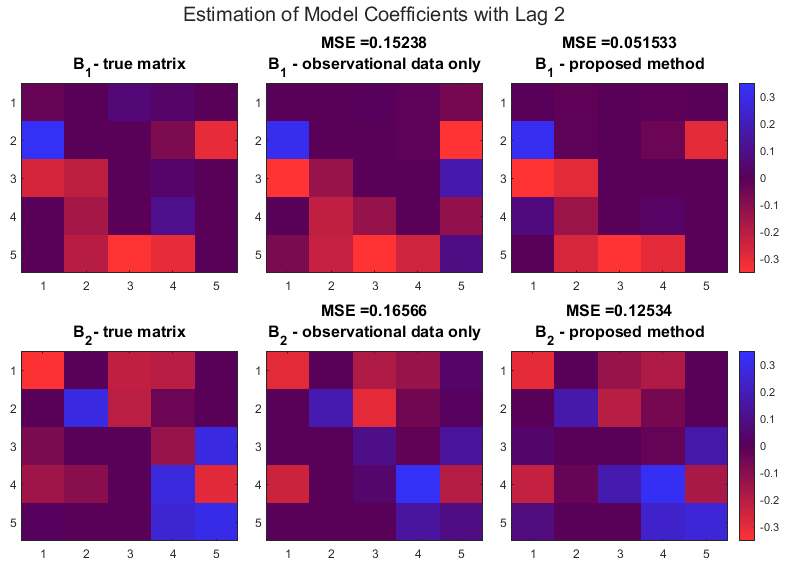}
    \caption{Learning the causal model of a small system. The performance is quantified by the MSE of the estimated entries of  ${\bf B}_q$, defined by $\text{MSE}=\frac{1}{D^2} \sum_{i,j} (b_{q,i,j}-\hat{b}_{q,i,j})^2$.}
    \label{fig: learning B}
\end{figure}

\subsection{Learning causal models with joint regression}

We simulated a time-series of length $T=390$, and system dimension $D=5$ with lag $Q=2$. Each element $b_{q,i,j}$ in the matrix coefficients $\B_q$ was generated as $b_{q,i,j} \sim \mathcal{U}(-0.5,0.5)$, for $q=1,2$. We set about $~30\%$ sparsity for each matrix. The dynamics were perturbed with Gaussian noise $\w_t \sim \mathcal{N}(\bzeros, \C)$, where the covariance $\C$ is a Toeplitz matrix with $C_{ii} = 1$, $C_{ij} = 0.5 \text{ if } |i-j|=1$, and $0$ otherwise. We intervened on the node $x_{1,\tau}$ within the interval $\tau \in \{101, 170\}$ and the node $x_{2,\tau}$ in the interval $\tau \in \{201, 270 \}$. For this example, we let the intervened nodes be driven by a sine wave, specifically $x_{i,\tau} := \pearl{u}_{i,t}$, where $\pearl{u}_{i,t}= 10\sin(\frac{\tau}{2})$, for $i=1,2$ in their respective time intervals. The rest of the nodes ($i=3,4,5$) were left unperturbed. We learned the $\B$ coefficients in two ways: i) by using only the observational (unintervened data ${\cal T}_0$), and ii) using intervened data as well - the proposed method. To insure a fair comparison, we opted to use the same number of data points $T_0=250$ for learning in both i) and ii), where in the case of i), $140$ of those points are the interventional dataset. With the current setup, the proposed method is at a slight disadvantage in regards to amount of data used, since we lose $140$ points for learning the coefficients $\b_{q,i}$ for $i=1,2$, $q=1,2$. However, those same $140$ points of intervention %
allow us to better estimate the coefficients $\b_{q,i}$ for $q=3,4,5$. %

In Figure \ref{fig: learning B}, we show the true matrices $\B_q$, for $q=1,2 $, their estimate using ${\cal T}_0$ only, and their estimate with the proposed method, i.e., using interventional data as well. We can see there is a drop in MSE when using the proposed method, which is also reflected visually in the heat maps. The effect becomes even more prominent when learning larger systems.

\subsection{Predicting the effects of a counterfactual intervention}
In this experiment, we consider the problem of predicting the effects of a counterfactual intervention in the past. %
Consider a multivariate time series of length $T=100$ and dimension $D=2$. We assume that the causal model is given by a VAR model, \eqref{eq:ARmodel}, with $Q=2$, whose parameters are known from prior investigation. Let us consider a counterfactual situation in which, at times $t=240$ until the end of the experiment, the experimenter intervenes upon $x_{1,t}$ by driving the signal with a new signal, $u(t)\sim \mathcal{N}(1,1)$.

\begin{figure}[h]
    \centering
    \includegraphics[width=0.45\textwidth]{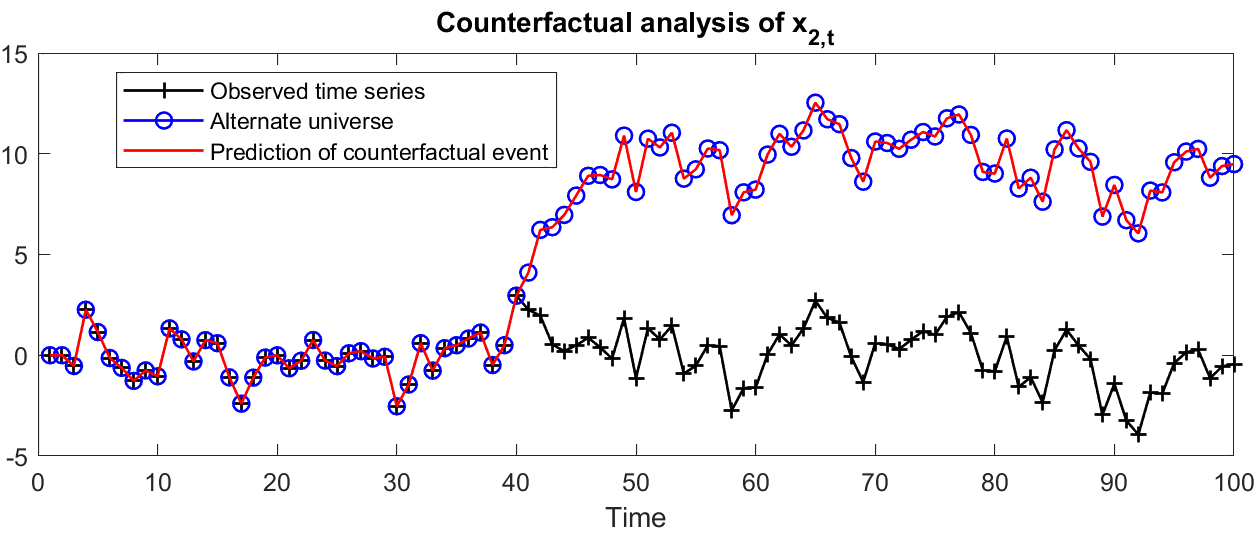}
    \caption{Predicting the effects of a past counterfactual intervention. We consider a counterfactual (hypothetical) intervention in which, starting at time 40. 
    We visualize three curves: the original observed signal $x_{2,t}$ (black), the signal which would have been observed if the hypothetical intervention upon  $x_{1,t}$ was performed (blue), and our prediction of what would have been observed in the counterfactual situation (red). 
    The black and blue curves were generated precisely by simulation of the system, and the red curve is produced by making predictions using the observed realization.}
    \label{fig:predcounterfactual}
    \vspace{-0.5cm}
\end{figure}

In Figure \ref{fig:predcounterfactual}, we predict the effects of the counterfactual intervention on $x_{1,t}$, as perceived by another variable $x_{2,t}$. 
We plot the signal $x_{2,t}$ that we observed originally alongside its variant in an ``alternate universe,'' in which the counterfactual intervention is performed. 
For times $t < 40$, both the observed time series and the counterfactual one coincide, as no intervention has not
occurred yet. For $40\leq t$, we see the counterfactual intervention begin to make the two time series diverge due to the effects of the intervention.  
Again, the model is able to well anticipate the counterfactual universe. 

\section{Conclusion}
In this work, we studied the problems of counterfactual reasoning and causal inference in the context of vector autoregressive models. 
By considering the system as modular and assigning separate functions to each variable, we can approach regression and model learning in a nonlinear manner.
Learning of the causal model is a necessary requirement to reason about counterfactual interventions and other what-if scenarios. 
The analysis of counterfactuals for nonlinear settings is analogous to what we have presented here, although additional care must be taken to estimate unobserved variables like the noises. Future work involves understanding the interventions for improved learning of the studied system and combining counterfactual analysis with interventions to achieve desired outcomes. Other directions include extending the counterfactual analysis to nonlinear models and models where the functional relationships are unknown and need to be estimated.

\bibliographystyle{ieeetr}
\bibliography{refs.bib}

\begin{thebibliography}{10}

\bibitem{pearl2009causality}
J.~Pearl, {\em Causality}.
\newblock Cambridge University Press, 2009.

\bibitem{rubin1974estimating}
D.~B. Rubin, ``Estimating causal effects of treatments in randomized and nonrandomized studies.,'' {\em Journal of Educational Psychology}, vol.~66, no.~5, p.~688, 1974.

\bibitem{spirtes2000causation}
P.~Spirtes, C.~N. Glymour, R.~Scheines, and D.~Heckerman, {\em Causation, Prediction, and Search}.
\newblock MIT Press, 2000.

\bibitem{zheng2018dags}
X.~Zheng, B.~Aragam, P.~K. Ravikumar, and E.~P. Xing, ``{DAG}s with {NO} {TEARS}: Continuous optimization for structure learning,'' {\em Advances in Neural Information Processing Systems}, vol.~31, 2018.

\bibitem{schreiber2000measuring}
T.~Schreiber, ``Measuring information transfer,'' {\em Physical Review Letters}, vol.~85, no.~2, p.~461, 2000.

\bibitem{peters2017elements}
J.~Peters, D.~Janzing, and B.~Sch{\"o}lkopf, {\em Elements of causal inference: foundations and learning algorithms}.
\newblock The MIT Press, 2017.

\bibitem{granger1969investigating}
C.~W. Granger, ``Investigating causal relations by econometric models and cross-spectral methods,'' {\em Econometrica: Journal of the Econometric Society}, pp.~424--438, 1969.

\bibitem{sugihara2012detecting}
G.~Sugihara, R.~May, H.~Ye, C.-h. Hsieh, E.~Deyle, M.~Fogarty, and S.~Munch, ``Detecting causality in complex ecosystems,'' {\em Science}, vol.~338, no.~6106, pp.~496--500, 2012.

\bibitem{butler2023causal}
K.~Butler, G.~Feng, and P.~M. Djurić, ``On causal discovery with convergent cross mapping,'' {\em IEEE Transactions on Signal Processing}, vol.~71, pp.~2595--2607, 2023.

\bibitem{ogarrio2016hybrid}
J.~M. Ogarrio, P.~Spirtes, and J.~Ramsey, ``A hybrid causal search algorithm for latent variable models,'' in {\em Conference on Probabilistic Graphical Models}, pp.~368--379, PMLR, 2016.

\bibitem{eberhardt2006n}
F.~Eberhardt, C.~Glymour, and R.~Scheines, ``{N-1} experiments suffice to determine the causal relations among {N} variables,'' {\em Innovations in Machine Learning: Theory and Applications}, pp.~97--112, 2006.

\bibitem{tibshirani1996regression}
R.~Tibshirani, ``Regression shrinkage and selection via the {L}asso,'' {\em Journal of the Royal Statistical Society Series B: Statistical Methodology}, vol.~58, no.~1, pp.~267--288, 1996.

\bibitem{cui2024topology}
C.~Cui, P.~Banelli, and P.~M. Djuri{\'c}, ``Topology inference of directed graphs by gaussian processes with sparsity constraints,'' {\em IEEE Transactions on Signal Processing}, 2024.

\bibitem{butler2022differential}
K.~Butler, G.~Feng, and P.~M. Djuri{\'c}, ``A differential measure of the strength of causation,'' {\em IEEE Signal Processing Letters}, vol.~29, pp.~2208--2212, 2022.

\bibitem{liu2023detecting}
Y.~Liu, C.~Cui, D.~Waxman, K.~Butler, and P.~M. Djuri{\'c}, ``Detecting confounders in multivariate time series using strength of causation,'' in {\em 2023 31st European Signal Processing Conference (EUSIPCO)}, pp.~1400--1404, IEEE, 2023.

\bibitem{salehkaleybar2020learning}
S.~Salehkaleybar, A.~Ghassami, N.~Kiyavash, and K.~Zhang, ``Learning linear non-{G}aussian causal models in the presence of latent variables,'' {\em Journal of Machine Learning Research}, vol.~21, no.~39, pp.~1--24, 2020.

\end{thebibliography}

\end{document}